\pdfoutput=1
\documentclass[11pt,logo,copyright]{nvidiatechreport}
\usepackage[authoryear,sort&compress,round]{natbib}
\renewcommand{\today}{}
\usepackage{amsmath,amssymb,amsthm,mathtools,bm}

\def\eqref#1{equation~\ref{#1}}

\def\1{\bm{1}}

\DeclareMathAlphabet{\mathsfit}{\encodingdefault}{\sfdefault}{m}{sl}
\SetMathAlphabet{\mathsfit}{bold}{\encodingdefault}{\sfdefault}{bx}{n}

\newcommand{\KL}{D_{\mathrm{KL}}}

\newcommand{\Cov}{\mathrm{Cov}}

\makeatletter
\let\save@mathaccent\mathaccent
\newcommand*\if@single[3]{%
  \setbox0\hbox{${\mathaccent"0362{#1}}^H$}%
  \setbox2\hbox{${\mathaccent"0362{\kern0pt#1}}^H$}%
  \ifdim\ht0=\ht2 #3\else #2\fi
  }
\newcommand*\rel@kern[1]{\kern#1\dimexpr\macc@kerna}
\newcommand*\widebar[1]{\@ifnextchar^{{\wide@bar{#1}{0}}}{\wide@bar{#1}{1}}}
\newcommand*\wide@bar[2]{\if@single{#1}{\wide@bar@{#1}{#2}{1}}{\wide@bar@{#1}{#2}{2}}}
\newcommand*\wide@bar@[3]{%
  \begingroup
  \def\mathaccent##1##2{%
    \let\mathaccent\save@mathaccent
    \if#32 \let\macc@nucleus\first@char \fi
    \setbox\z@\hbox{$\macc@style{\macc@nucleus}_{}$}%
    \setbox\tw@\hbox{$\macc@style{\macc@nucleus}{}_{}$}%
    \dimen@\wd\tw@
    \advance\dimen@-\wd\z@
    \divide\dimen@ 3
    \@tempdima\wd\tw@
    \advance\@tempdima-\scriptspace
    \divide\@tempdima 10
    \advance\dimen@-\@tempdima
    \ifdim\dimen@>\z@ \dimen@0pt\fi
    \rel@kern{0.6}\kern-\dimen@
    \if#31
      \overline{\rel@kern{-0.6}\kern\dimen@\macc@nucleus\rel@kern{0.4}\kern\dimen@}%
      \advance\dimen@0.4\dimexpr\macc@kerna
      \let\final@kern#2%
      \ifdim\dimen@<\z@ \let\final@kern1\fi
      \if\final@kern1 \kern-\dimen@\fi
    \else
      \overline{\rel@kern{-0.6}\kern\dimen@#1}%
    \fi
  }%
  \macc@depth\@ne
  \let\math@bgroup\@empty \let\math@egroup\macc@set@skewchar
  \mathsurround\z@ \frozen@everymath{\mathgroup\macc@group\relax}%
  \macc@set@skewchar\relax
  \let\mathaccentV\macc@nested@a
  \if#31
    \macc@nested@a\relax111{#1}%
  \else
    \def\gobble@till@marker##1\endmarker{}%
    \futurelet\first@char\gobble@till@marker#1\endmarker
    \ifcat\noexpand\first@char A\else
      \def\first@char{}%
    \fi
    \macc@nested@a\relax111{\first@char}%
  \fi
  \endgroup
}
\makeatother

\usepackage{graphicx}
\usepackage{array,booktabs,longtable,tabularx}
\usepackage{algorithm}
\usepackage[noend]{algpseudocode}
\usepackage{enumitem}
\usepackage{microtype}
\usepackage{xurl}
\usepackage{xspace}
\usepackage{xcolor}
\usepackage{float}
\AddToHook{env/table/begin}{%
  \setlength{\abovecaptionskip}{2pt}%
  \setlength{\belowcaptionskip}{4pt}}
\usepackage{placeins}
\usepackage{tikz}
\usepackage{pgfplots}
\usepgfplotslibrary{groupplots}
\usetikzlibrary{arrows.meta,positioning,calc,fit,backgrounds}
\pgfplotsset{compat=1.18}
\usepackage{hyperref}
\usepackage{url}
\usepackage[nameinlink,noabbrev]{cleveref}

\definecolor{appleBlue}{HTML}{007AFF}
\definecolor{appleGreen}{HTML}{34C759}
\definecolor{appleIndigo}{HTML}{5856D6}
\definecolor{appleOrange}{HTML}{FF9500}
\definecolor{appleRed}{HTML}{FF3B30}
\definecolor{applePurple}{HTML}{AF52DE}
\definecolor{appleTeal}{HTML}{5AC8FA}
\definecolor{appleGray}{HTML}{8E8E93}
\definecolor{appleGray2}{HTML}{AEAEB2}
\definecolor{appleGray5}{HTML}{E5E5EA}
\definecolor{appleGray6}{HTML}{F2F2F7}
\definecolor{appleInk}{HTML}{1C1C1E}

\definecolor{nvGreen}{HTML}{76B900}
\definecolor{nvGreenDark}{HTML}{3F7A00}
\definecolor{nvTeal}{HTML}{0F8B8D}
\definecolor{nvSlate}{HTML}{2F7DE1}
\definecolor{nvAmber}{HTML}{E08A00}
\definecolor{nvRed}{HTML}{C0392B}
\definecolor{nvInk}{HTML}{1D1D1F}
\definecolor{nvGray}{HTML}{666666}
\definecolor{nvGray2}{HTML}{D2D2D7}
\definecolor{nvGray6}{HTML}{F5F5F7}

\hypersetup{
  colorlinks=true,
  linkcolor=appleBlue,
  citecolor=appleIndigo,
  urlcolor=appleBlue
}

\tikzset{
  >={Stealth[length=2.2mm]},
  applebox/.style={
    rounded corners=3pt,
    line width=0.8pt,
    inner sep=6pt,
    align=center,
    font=\small,
    text=appleInk
  },
  actorbox/.style={applebox,draw=appleBlue,fill=appleBlue!7},
  recordbox/.style={applebox,draw=appleOrange,fill=appleOrange!8},
  learnerbox/.style={applebox,draw=appleGreen,fill=appleGreen!8},
  controllerbox/.style={applebox,draw=appleIndigo,fill=appleIndigo!8},
  neutralbox/.style={applebox,draw=appleGray2,fill=appleGray6},
  nvbox/.style={applebox,rounded corners=5pt,text=nvInk},
  nvneutralbox/.style={nvbox,draw=nvGray2,fill=nvGray6},
  flowarrow/.style={-{Stealth},line width=0.9pt,draw=appleGray},
  flowlabel/.style={font=\footnotesize,text=appleInk,fill=white,inner sep=2pt,align=center},
  lanelabel/.style={font=\small\bfseries,anchor=east,align=right},
  laneband/.style={rounded corners=2pt},
  timedot/.style={circle,inner sep=0pt,minimum size=4.4pt},
  dimline/.style={{Stealth[length=2mm]}-{Stealth[length=2mm]},line width=0.9pt},
  dimtag/.style={font=\footnotesize,inner sep=2.5pt},
  dotnote/.style={font=\footnotesize,text=appleInk,inner sep=1.5pt}
}

\newsavebox{\fitbox}
\newcommand{\fitfigure}[1]{%
  \sbox{\fitbox}{#1}%
  \ifdim\wd\fitbox>\columnwidth
    \resizebox{\columnwidth}{!}{\usebox{\fitbox}}%
  \else
    \usebox{\fitbox}%
  \fi}

\newsavebox{\tablebox}
\newcommand{\fittable}[2][\columnwidth]{%
  \sbox{\tablebox}{#2}%
  \ifdim\wd\tablebox>#1
    \resizebox{#1}{!}{\usebox{\tablebox}}%
  \else
    \usebox{\tablebox}%
  \fi}

\newtheorem{thm}{Theorem}[section]
\newtheorem{prop}[thm]{Proposition}

\theoremstyle{definition}

\theoremstyle{remark}

\crefname{equation}{Eq.}{Eqs.}
\Crefname{equation}{Equation}{Equations}
\crefname{figure}{Fig.}{Figs.}
\Crefname{figure}{Figure}{Figures}
\crefname{section}{Sec.}{Secs.}
\Crefname{section}{Section}{Sections}
\crefname{table}{Tab.}{Tabs.}
\Crefname{table}{Table}{Tables}
\crefname{algorithm}{Alg.}{Algs.}
\Crefname{algorithm}{Algorithm}{Algorithms}
\crefname{thm}{Thm.}{Thms.}
\Crefname{thm}{Theorem}{Theorems}
\crefname{prop}{Prop.}{Props.}
\Crefname{prop}{Proposition}{Propositions}
\crefname{lemma}{Lemma}{Lemmas}
\crefname{cor}{Cor.}{Cors.}
\crefname{definition}{Def.}{Defs.}
\crefname{assumption}{Assumption}{Assumptions}
\crefname{remark}{Remark}{Remarks}

\providecommand{\KL}{D_{\mathrm{KL}}}
\newcommand{\TV}{D_{\mathrm{TV}}}

\newcommand{\FR}{\textsc{FlashREINFORCE}\xspace}

\newcommand{\ind}{\mathbf{1}}
\DeclareMathOperator{\sg}{stopgrad}

\newcommand{\FRtableformat}{%
  \renewcommand{\arraystretch}{1.12}}
\newcommand{\FRtable}{\small\setlength{\tabcolsep}{5pt}\FRtableformat}

\newif\ifFRsubmissionstatements
\FRsubmissionstatementsfalse

\newcommand{\FRCodeAvailabilityNote}{%
  \footnotetext[1]{Code and configurations will be released after the anonymous review period.}}
\newcommand{\FRCodeAvailabilityStatement}{%
  Code and configurations will be released after the anonymous review period, as stated on the first page.}

\title{FlashREINFORCE: Critic-Free Single-Rollout Asynchronous RL for Agentic Language Models}

\newcommand{\FRCodeLink}{\url{https://github.com/NVIDIA-NeMo/labs-molt}}
\renewcommand{\FRCodeAvailabilityStatement}{Code: \FRCodeLink.}

\renewcommand{\FRCodeAvailabilityNote}{%
  \begingroup
  \renewcommand{\thefootnote}{}%
  \footnotetext{\FRCodeAvailabilityStatement}%
  \endgroup
}

\author[1]{Jian Hu$^{*,\dagger}$}
\author[1]{Yifan Zhang$^{*}$}
\author[1]{Hao Zhang$^{*}$}
\author[1]{Binfeng Xu$^{*}$}
\author[1]{Shaokun Zhang}
\author[1]{Hongqing Peng}
\author[1]{Zhiding Yu}
\author[1]{Pavlo Molchanov}
\author[1]{Jan Kautz}
\author[1]{Yi Dong$^{\dagger}$}
\affil[1]{NVIDIA}

\AddToHook{cmd/maketitle/after}{%
  \begingroup
  \renewcommand{\thefootnote}{\fnsymbol{footnote}}%
  \footnotetext[1]{Equal contribution.}%
  \footnotetext[2]{This research was led by Yi Dong and Jian Hu.
    Contact: \href{mailto:yidong@nvidia.com}{yidong@nvidia.com},
    \href{mailto:jianh@nvidia.com}{jianh@nvidia.com}.}%
  \endgroup
}

\hypersetup{
  pdftitle={FlashREINFORCE: Critic-Free Single-Rollout Asynchronous RL for Agentic Language Models},
  pdfauthor={Jian Hu, Yifan Zhang, Hao Zhang, Binfeng Xu, Shaokun Zhang, Hongqing Peng, Zhiding Yu, Pavlo Molchanov, Jan Kautz, Yi Dong},
  pdfsubject={NVIDIA Technical Report}
}

\begin{document}
\maketitle

\begin{abstract}
Asynchronous reinforcement learning suits long-horizon agents with irregular rollout times, but group-relative updates require repeated rollouts of each prompt.  However, group-based sampling limits prompt coverage at a fixed rollout budget and introduces group synchronization barriers.  One rollout per prompt avoids these constraints but requires learning from stale, variable-length trajectories without a sibling baseline.  We present \FR, to our knowledge the first critic-free single-rollout framework that scales to long asynchronous training and large models.  It combines One-Batch REINFORCE, a Sequence Trust Region, and Sample-Mean Optimization, which provide prompt coverage and batch centering while controlling policy drift and length-amplified negative updates.  Each component is easy to implement, and the result is their combination: a batch-mean baseline is stable under sample-mean loss, and each common alternative degrades or collapses training.  \FR remains stable through 6,000 updates on DeepSeek-R1-Distill-Qwen-1.5B at policy lag four, and through 1,450 on the 30B MoE Qwen3-30B-A3B at lag eight.  Under identical loss, data, and lag it matches a strengthened group-relative baseline within 0.5 points, and with half the rollouts exceeds published synchronous baselines on Qwen2.5-Math-1.5B (38.0 versus 36.3 five-benchmark mean).  On Qwen2.5-7B-Instruct it maintains tool use where GRPO stops calling tools and achieves 98.3\%/96.5\% seen/unseen ALFWorld success.
\end{abstract}
\abscontent

\FRCodeAvailabilityNote

\section{Introduction}
\label{sec:introduction}

In long-horizon agent RL, tool calls, environment steps, and variable-length reasoning make rollout durations unpredictable \citep{noukhovitch2025asyncrlhf,fu2025areal,wu2025llamarl}.  Asynchronous workers can submit each completed trajectory immediately, but methods such as GRPO estimate advantages from groups of sibling rollouts \citep{shao2024deepseekmath}.  However, group-based sampling limits prompt coverage at a fixed rollout budget, introduces group synchronization barriers, and is difficult to apply to online interactions that yield only a single trajectory.  A budget of $G$ rollouts could instead cover $G$ distinct prompts, with each trajectory available for learning as soon as it finishes.

\begin{table}[!htb]
\caption{Requirements of critic-free and single-rollout methods.  \FR is critic-free, uses one rollout per prompt, and trains asynchronously.}
\label{tab:positioning}
\centering
\FRtable
\begin{tabular}{@{}lccc@{}}
\toprule
\textbf{Method} & \textbf{No} & \textbf{One rollout} & \textbf{Asynchronous}\\
 & \textbf{critic} & \textbf{per prompt} & \textbf{training}\\
\midrule
GRPO \citep{shao2024deepseekmath} & \checkmark & -- & group wait\\
SAO \citep{hou2026sao} & -- & \checkmark & \checkmark\\
C-RF + NTF \citep{wu2026crf} & \checkmark & \checkmark & --\\
\textbf{\FR} & \checkmark & \checkmark & \checkmark\\
\bottomrule
\end{tabular}
\end{table}

The difficulty is learning reliably from one rollout per prompt.  We instead center rewards across prompts.  With binary rewards, positive-only feedback gives failures no update; batch centering allows them to contribute, but applies a coarse penalty to the whole trajectory.  A failed trajectory can contain useful intermediate steps, yet its negative advantage penalizes every token.  Length weighting then gives long failures extra influence.  The challenge is to retain negative feedback without letting trajectory length amplify it.

Asynchrony adds a second challenge: completed trajectories may come from different policy snapshots, leaving both action-probability and history-distribution mismatch with the learner.  Token importance sampling (IS) corrects actions at stored histories, but those histories still follow the behavior policy.  Token correction therefore leaves a reason to consider drift over the whole trajectory.  Our analysis relates the remaining surrogate error to accumulated policy drift, motivating sequence-level trust alongside IS.

Prior single-rollout methods either learn a value model, as in SAO \citep{hou2026sao}, or are demonstrated with synchronous training, as in Contrastive-REINFORCE with Negative Token Filtering (C-RF + NTF; \citealp{wu2026crf}) (\cref{tab:positioning}).  We ask whether critic-free single-rollout training can scale to long asynchronous runs and large models.

\begin{figure}[!htb]
\centering
\includegraphics[width=0.62\linewidth]{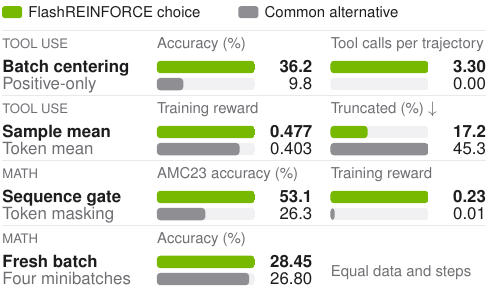}
\caption{Each design choice of \FR (green) and the result of switching it alone to its common alternative (gray; \cref{sec:ablations}).  Accuracy is the three-benchmark mean unless named; $\downarrow$: lower is better.  Token mean is shown 150 updates after the switch: its reward has barely moved, but 45.3\% truncation leaves the run untrainable.  Token masking collapses near update 1,200.  Tool use: Qwen2.5-7B-Instruct; math: Qwen2.5-Math-1.5B.}
\label{fig:choices}
\end{figure}

\FR combines three complementary components.  \emph{One-Batch REINFORCE} applies batch centering across independent prompts.  The \emph{Sequence Trust Region} complements token IS with trajectory-level drift screening.  \emph{Sample-Mean Optimization} normalizes trajectory contributions to limit length-amplified negative feedback.  Together, these components enable critic-free, one-pass asynchronous updates without waiting for sibling rollouts.  To our knowledge, \FR is the first critic-free single-rollout method shown to scale in training length, staleness, and model size.

The contribution is this recipe and the scale at which it holds.  Every choice is easy to implement, and that is its strength: a good recipe needs few code changes and delivers strong practical results.  This one also simplifies the pipeline, with no group size, clipping range, or critic to set.  What makes the setting work is the combination, because the choices depend on one another.  Centering across prompts gives failed trajectories a negative advantage, which is why the loss averages within each trajectory: a batch mean, reported as unstable \citep{liu2025tricks}, collapses under token-mean loss and trains stably under sample mean (\cref{sec:objective}).  Asynchrony makes stored histories stale, which is why admission acts on whole trajectories.  Making any one choice the common way fails in its own way: tool calls stop, responses run to the length limit, reward collapses, or accuracy drops (\cref{fig:choices}).  Scale is what exposes these failures: token-level masking collapses near update 1,200, and without the gate accuracy plateaus after about 4,000 updates while the gated run keeps improving (\cref{sec:ablations}), so a short run cannot tell a working recipe from a failing one.  TD3 and Rainbow are results of the same kind: combinations of simple changes, valued for the setting they made work \citep{fujimoto2018td3,hessel2018rainbow}.

\paragraph{Contributions.}
\begin{itemize}[leftmargin=*,nosep]
  \item \textbf{Critic-free single-rollout RL scales.}  \FR trains stably through 6,000 asynchronous updates at policy lag $\approx4$, raising AIME24/25 mean accuracy from 21.7 to 33.7, and through 1,450 updates on a 30B MoE at lag eight (\cref{sec:r1-main,sec:tool-main}).  On agent tasks it sustains tool use where GRPO stops calling tools and reaches 98.3\%/96.5\% seen/unseen ALFWorld success.
  \item \textbf{Group-baseline accuracy at lower cost.}  Under identical loss, data, and lag, one rollout per prompt stays within 0.5 points of a Dr.\ GRPO-style group baseline (\cref{sec:group-batch-ablation}).  With half the rollouts, \FR has the highest five-benchmark mean on Qwen2.5-Math-1.5B against published synchronous baselines (\cref{sec:math-main}).
  \item \textbf{One hyperparameter, and a reason for each choice.}  The recipe uses one rollout per prompt, no critic, and one pass over each batch, and adds a single threshold $\delta$ (\cref{alg:flash}).  A reward--length covariance explains why a batch mean needs sample mean (\cref{sec:objective}), a drift bound motivates the sequence gate (\cref{sec:local-surrogate}), and each common alternative degrades or collapses training (\cref{fig:choices}).
\end{itemize}

\begin{figure}[!htb]
\centering
\linespread{1}\selectfont
\resizebox{\textwidth}{!}{%
\begin{tikzpicture}
  \def\cardH{3.5}
  \tikzset{
    card/.style={rounded corners=5pt,line width=0.8pt},
    cardtext/.style={align=center,inner sep=0pt,text=nvInk,font=\fontsize{8.5}{10.5}\selectfont},
    cardnote/.style={cardtext,font=\fontsize{7.5}{9.4}\selectfont}
  }
  \path[card,draw=nvGreen,fill=nvGreen!12] (0,0) rectangle (4.6,-\cardH);
  \node[cardtext,anchor=north,text width=4.3cm] (t1) at (2.3,-0.22)
    {\textcolor{nvGreenDark}{\textbf{1. One-batch update with mean baseline}}};
  \node[cardnote,anchor=south,text width=4.2cm] (n1) at (2.3,-\cardH+0.22)
    {Center one fresh batch; update once; discard.  No group wait or replay.};
  \node[cardtext,text width=4.2cm] at ($(t1.south)!0.5!(n1.north)$)
    {$\bar R=B^{-1}\sum_{j=1}^{B}R_j$\\[0.6mm] $A_i=R_i-\bar R$};
  \path[card,draw=nvSlate,fill=nvSlate!9] (4.9,0) rectangle (9.5,-\cardH);
  \node[cardtext,anchor=north,text width=4.2cm] (t2) at (7.2,-0.22)
    {\textcolor{nvSlate}{\textbf{2. Sequence trust region}}};
  \node[cardnote,anchor=south,text width=4.2cm] (n2) at (7.2,-\cardH+0.22)
    {Accept only if the mean sampled-action proxy is at or below the sequence threshold.};
  \node[cardtext] at ($(t2.south)!0.5!(n2.north)$)
    {$m_i=\ind[\bar D_i\leq\delta]$};
  \path[card,draw=nvGray2,fill=nvGray6] (9.8,0) rectangle (17.2,-\cardH);
  \node[cardtext,anchor=north,text width=7.0cm] (t3) at (13.5,-0.22)
    {\textbf{3. Sample-mean loss instead of token-mean loss}};
  \node[cardnote,anchor=south,text width=7.0cm] (n3) at (13.5,-\cardH+0.22)
    {\textcolor{nvGreenDark}{$A_i$}: centered advantage; \textcolor{nvAmber}{$\pi_\theta/\mu_i$}: learner-to-behavior ratio; $T_i^{-1}$: within-trajectory average; \textcolor{nvSlate}{$m_i$}: one admission decision for the complete trajectory.};
  \node[cardtext] at ($(t3.south)!0.5!(n3.north)$)
    {$\displaystyle
    \widehat{\mathcal J}_{\mathrm{Flash}}(\theta)
    =\frac1B\sum_{i=1}^{B}\frac{
      \textcolor{nvSlate}{m_i}\,
      \textcolor{nvGreenDark}{A_i}}{T_i}
      \sum_{t=1}^{T_i}
      \textcolor{nvAmber}{\frac{\pi_\theta(a_{i,t}\mid h_{i,t})}{\mu_i(a_{i,t}\mid h_{i,t})}}$};
\end{tikzpicture}%
}
\caption{\textbf{FlashREINFORCE at a glance:} one fresh-batch update, whole-trajectory admission, sample-mean loss.}
\label{fig:method-overview}
\end{figure}

\section{Setup}
\label{sec:setup}

\subsection{Asynchronous Rollout Data}

Rollout workers periodically receive policy snapshots, sample one trajectory per prompt, and submit trajectories independently as they finish.  The learner forms each batch from the next $B$ completions, which may come from several snapshots; the snapshot that generated a trajectory is its \emph{behavior policy}.  A batch is \emph{fresh} if none of its trajectories has been used for an update; its trajectories can still be stale, that is, generated by older snapshots.  \emph{Policy lag} is the age of the behavior policy in learner updates, and \emph{policy drift} is how far the learner's token distributions are from it.

For prompt $x_i$, let $\tau_i$ contain policy-visible histories $h_{i,t}$ (the context before each token), generated tokens $a_{i,t}$ for $t=1,\ldots,T_i$, and a scalar terminal reward $R_i\in\mathbb R$.  Let $\mu_i(a_{i,t}\mid h_{i,t})$ denote the behavior probability that actually generated token $a_{i,t}$; we store it with the trajectory.  Recomputing an ``old'' probability later need not recover the sampling probability because the learner and inference engine may already differ \citep{zhang2026prls1,zhong2026tim,zhang2025tbik}.  At learner parameters $\theta$, the token importance ratio is
\begin{equation}
 \rho_{i,t}(\theta)
 =\frac{\pi_\theta(a_{i,t}\mid h_{i,t})}{\mu_i(a_{i,t}\mid h_{i,t})}.
 \label{eq:ratio}
\end{equation}

\subsection{What Token Importance Sampling Corrects}
\label{sec:local-surrogate}

Let $J(\pi)=\mathbb E_{\tau\sim\pi}[R(\tau)]$ be the expected return and $\mu$ a behavior policy.  Let $\mathcal L_\mu(\pi)$ be the local surrogate that evaluates the actions of $\pi$ on histories and continuation values generated by $\mu$ (\cref{eq:local-surrogate}, \cref{app:granularity-theory}).  Its gradient is the token-IS estimator,
\begin{equation}
 \nabla_\theta\mathcal L_\mu(\pi_\theta)
 =\mathbb E_{\tau\sim\mu}\Big[\sum_t\rho_t\,R(\tau)\,\nabla_\theta\log\pi_\theta(a_t\mid h_t)\Big]
 \label{eq:local-gradient}
\end{equation}
(\cref{prop:local-gradient}).  Token IS thus corrects the action distribution at every stored history.  The histories themselves still follow $\mu$, whereas $J(\pi_\theta)$ depends on histories generated by $\pi_\theta$.  For batches with several behavior snapshots, the identity applies to each trajectory's $\mu_i$ before averaging.

Full-sequence or prefix ratios can correct the histories as well, but multiplying ratios across a long trajectory can produce high variance.  We keep the token ratio and bound the remaining error.  For a fixed horizon $T$, let $\varepsilon_t$ and $\kappa_t$ be the largest total-variation and KL divergences between $\pi$ and $\mu$ at position $t$, let $\bar\kappa$ be the mean of $\kappa_t$ over positions, and let $A_{\max}$ bound the behavior advantage.  Then
\begin{equation}
 |J(\pi)-\mathcal L_\mu(\pi)|
 \leq 2A_{\max}\Big(\sum_t\varepsilon_t\Big)^2
 \leq A_{\max}T^2\bar\kappa
 \label{eq:accumulated-drift-main}
\end{equation}
(\cref{prop:accumulated-drift}), by the classical performance-difference and coupling argument \citep{kakade2002cpi,schulman2015trpo}.  The error grows with drift summed over the whole trajectory.  This motivates accepting or rejecting whole trajectories: masking individual tokens leaves the contexts of the remaining tokens unchanged.

\section{FlashREINFORCE}
\label{sec:method}

\FR combines fresh-batch reward centering, sequence-level trust, and trajectory-normalized updates (\cref{fig:method-overview,alg:flash}).  \Cref{alg:flash} is the complete update; implementing it needs none of the analysis.

\subsection{One-Batch REINFORCE with Reward Centering}
\label{sec:advantage}

With one rollout per prompt, a sibling-rollout baseline is unavailable; with binary rewards, uncentered REINFORCE also gives failures no learning signal.  \FR therefore applies batch centering over the next $B$ completed trajectories, using the batch-level normalization of REINFORCE++ \citep{hu2025reinforcepp} in mean-only form, to balance positive and negative gradients without a critic,
\begin{equation}
 \bar R=\frac1B\sum_{j=1}^{B}R_j,
 \qquad
 A_i=R_i-\bar R.
 \label{eq:baseline}
\end{equation}
The advantages sum to zero, so positive ($A_i>0$) and negative ($A_i<0$) samples carry equal total weight in the update at any success rate.  Each fresh batch receives one full-batch update and is discarded, avoiding additional drift from sequential updates on the same collected batch.

At a fixed rollout budget, one rollout per prompt covers $G$ times as many prompts as groups of $G$, while a group mean adapts the baseline to each prompt.  \citet{hu2026groupbatch} analyzes this tradeoff for on-policy leave-one-out centering, and our controlled $128\times1$ versus $32\times4$ comparison learns comparably for about 4,000 updates (\cref{sec:group-batch-ablation}).

\subsection{Sequence Trust Region}
\label{sec:sequence-trust-region}

The accumulated-drift bound in \cref{eq:accumulated-drift-main} motivates screening complete trajectories, complementing token IS with sequence-level control.  Earlier sequence-level masks threshold other statistics: Seq-MIS the product of token ratios over the response \citep{liu2025seqmis}, and Trust Region Masking the largest full-vocabulary token KL \citep{li2026trustregionmasking}.  Both statistics scale with trajectory length: a log-product sums over tokens, and a maximum is set by the single most-drifted token.  Ratio thresholds also measure relative change: as DPPO explains, they can over-restrict low-probability token updates that move little probability mass, yet admit larger mass shifts on high-probability tokens \citep[][Sec.~4.2]{qi2026dppo}, the asymmetry behind DAPO's clip-higher \citep{yu2025dapo}.  In our earlier runs, bounds of $0.999$ and $1.001$ on the geometric-mean ratio rejected up to 99\% of the data and training collapsed.  \FR instead gates on the trajectory mean of a sampled-action KL proxy, a computable counterpart of $\bar\kappa$ in \cref{eq:accumulated-drift-main} with one scale at every trajectory length.

In words, the gate measures how far the learner's probability of each sampled token has moved since generation, averages this over the trajectory, and drops the trajectory when the average exceeds a threshold $\delta$.  We use gate, admission, and trust region for this one decision.

For each sampled token, we collapse both distributions to the sampled action versus the remaining vocabulary mass, giving the Bernoulli KL proxy also studied in DPPO \citep[][Sec.~4.4]{qi2026dppo},
\begin{equation}
 d_{i,t}=p_{i,t}\log\frac{p_{i,t}}{q_{i,t}}
 +(1-p_{i,t})\log\frac{1-p_{i,t}}{1-q_{i,t}},
 \label{eq:sampled-action-kl}
\end{equation}
where $p_{i,t}=\mu_i(a_{i,t}\mid h_{i,t})$ and $q_{i,t}=\pi_\theta(a_{i,t}\mid h_{i,t})$.  This reward-independent proxy is a lower bound on categorical KL by the data-processing inequality and uses only the behavior probability already stored for IS.  The gate thresholds its mean over the trajectory,
\begin{equation}
 \bar D_i=\frac1{T_i}\sum_{t=1}^{T_i}d_{i,t},
 \qquad
 m_i=\ind[\bar D_i\leq\delta],
 \label{eq:sequence-trust-region}
\end{equation}
and $m_i$ is held fixed during the update.  Rejection removes a trajectory's complete contribution when its proxy exceeds $\delta$, which also implies mean full KL above $\delta$ (\cref{app:granularity-theory}).  As a general default we recommend $\delta=0.05$, which we adopted in preliminary OSWorld experiments with Nemotron 3 Nano Omni and have since found reliable in most settings we tried; the experiments below use tighter thresholds chosen for their settings ($10^{-3}$ to $10^{-2}$).

\subsection{Sample-Mean Optimization}
\label{sec:objective}

The sequence gate decides which trajectories are used; the loss reduction decides how much weight each one receives.  A negative advantage penalizes every token of a failed rollout, and token-mean reduction gives longer failures extra weight.  \FR instead averages within each trajectory before averaging across trajectories, retaining batch centering while removing the explicit length multiplier:
\begin{equation}
 \widehat{\mathcal J}(\theta)
 =\frac1B\sum_{i=1}^{B}\frac{m_iA_i}{T_i}
 \sum_{t=1}^{T_i}\rho_{i,t}(\theta).
 \label{eq:flash-objective}
\end{equation}
In practice we minimize the loss
\begin{equation}
 \mathcal L(\theta)
 =-\frac1B\sum_{i=1}^{B}\frac{m_iA_i}{T_i}
 \sum_{t=1}^{T_i}\sg(\rho_{i,t})\log\pi_\theta(a_{i,t}\mid h_{i,t}),
 \label{eq:flash-loss}
\end{equation}
where $\sg$ is the stop-gradient operator and $A_i$ and $m_i$ are held fixed.  Its negative gradient equals the gradient of \cref{eq:flash-objective} (\cref{eq:flash-gradient} in \cref{app:loss-reductions}).  The update uses no PPO-style ratio clipping, critic, or reference-model forward pass.  Sample mean gives each trajectory weight $1/B$, whereas token mean weights it by $T_i/\sum_jT_j$.  \Cref{app:token-mask} examines an optional filter on the tokens of failed trajectories.

\begin{algorithm}[!htb]
\caption{The one-pass \FR update.}
\label{alg:flash}
\small
\begin{algorithmic}
\Require completed trajectories $\mathcal B$, policy $\pi_\theta$, threshold $\delta$
\State \textcolor{nvGreenDark}{\textbf{1. Batch:}} $A_i\gets R_i-B^{-1}\sum_jR_j$
\State \textcolor{nvSlate}{\textbf{2. Trust:}} $\rho_{i,t}\gets\pi_\theta(a_{i,t}\mid h_{i,t})/\mu_i(a_{i,t}\mid h_{i,t})$
\State \hspace{1.05cm} $\bar D_i\gets T_i^{-1}\sum_td_{i,t}$;\quad $m_i\gets\ind[\bar D_i\leq\delta]$
\State \textcolor{nvInk}{\textbf{3. Update:}} one optimizer step on the loss in \cref{eq:flash-loss}, with $A_i$, $m_i$, and $\mu_i$ held fixed
\end{algorithmic}
\end{algorithm}

A batch mean needs this reduction to balance the token-gradient weight of positive and negative samples.  Before gating, sample-mean weights $A_i/B$ sum to zero (\cref{sec:advantage}); token-mean weights sum to $\Cov(R,T)/\bar T$, the reward--length covariance over the mean length (\cref{app:loss-reductions}).  When failures run longer than successes, negative samples outweigh positive ones and responses lengthen until 45.3\% are truncated (\cref{sec:reduction-ablation}).  GRPO-style methods may be balanced automatically: when responses to one prompt have similar lengths, a group-mean baseline cancels within each group.

\section{Experiments}
\label{sec:experiments}

We test long-run stability (\cref{sec:r1-main}), scale (\cref{sec:tool-main}), accuracy against a controlled group baseline (\cref{sec:group-batch-ablation}) and published baselines (\cref{sec:math-main}), and each design choice (\cref{sec:ablations}).

\paragraph{Training.}
All \FR runs collect rollouts asynchronously (\cref{sec:setup}) and apply the one-pass update of \cref{alg:flash}, with one rollout per prompt, no critic, and no reference model.  Policy lag is $\approx4$ for long-CoT reasoning and $\approx8$ for the 30B MoE.

\paragraph{Baselines.}
\label{sec:protocols}
We compare with two kinds of baselines.  The first are trained in our framework with the same rollouts per update as \FR: GRPO, and a strengthened group-mean variant that receives the full \FR loss.  The second are the published GRPO and C-RF + NTF results of \citet{wu2026crf}; C-RF + NTF contrasts successful and failed trajectories and masks high-probability tokens in failures.  These baselines were trained \emph{synchronously} with veRL \citep{sheng2024hybridflow}: each iteration samples all of its rollouts from the latest policy, so there is no policy lag.  \FR is trained \emph{asynchronously}, on rollouts several updates stale, so the setting gives it no advantage.

\paragraph{Evaluation.}
All accuracies are avg@32, the mean accuracy over 32 sampled responses per problem, with equally weighted benchmark means computed before rounding.  Datasets, sampling settings, and hyperparameters are in \cref{app:experimental-details}.  Code and configurations are available at \FRCodeLink.

\begin{figure}[!htb]
\centering
\includegraphics[width=\textwidth]{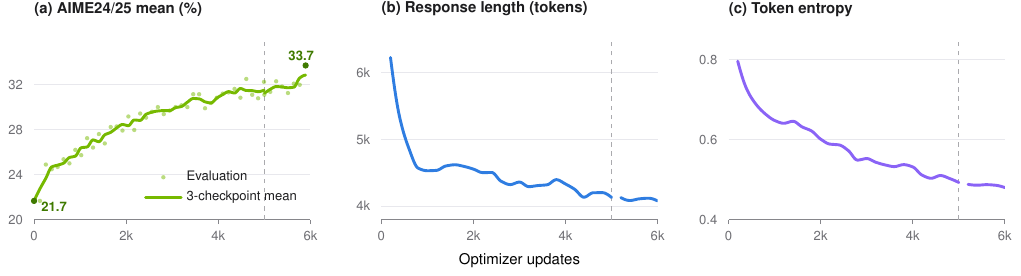}
\caption{DeepSeek-R1-Distill-Qwen-1.5B with sequence trust at policy lag $\approx4$, through 6,000 cumulative updates.  The dashed line marks continuation from the step-5,000 checkpoint (\cref{tab:r1-aime-scores}).  (a) AIME24/25 mean avg@32 accuracy: dots mark evaluations and curves show a three-checkpoint moving average within each segment.  (b,c) Training response length and token entropy, averaged in consecutive 200-update windows.}
\label{fig:r1-trust-stability}
\end{figure}

\subsection{Sustained Asynchronous Training}
\label{sec:r1-main}

\FR remains stable through 6,000 cumulative updates at policy lag $\approx4$ on DeepSeek-R1-Distill-Qwen-1.5B, improving its AIME24/25 mean from 21.7 to 33.7 (\cref{fig:r1-trust-stability}).  Responses shorten and entropy settles over training.

\subsection{Tool Use and a 30B MoE at Lag Eight}
\label{sec:tool-main}
\label{sec:moe-tool-results}

\begin{table}[!htb]
\caption{Python-tool avg@32 accuracy (\%) at matched steps and rollout budgets.  Calls per trajectory are averaged over the 100 updates ending at each checkpoint.  In (b), \FR uses $\delta=10^{-2}$ at lag $\approx8$ and GRPO uses lag $\approx1$.}
\label{tab:tool-main}
\centering
\FRtable
\begin{tabular}{@{}lrrrrr@{}}
\toprule
\textbf{Method} & \textbf{AMC23} & \textbf{Minerva} & \textbf{AIME25} & \textbf{Mean} & \textbf{Calls}\\
\midrule
\multicolumn{6}{@{}l}{\textit{(a) Qwen2.5-7B-Instruct, step 600}}\\
GRPO & 51.2 & \textbf{32.3} & 7.5 & 30.3 & 0.00\\
\FR & \textbf{60.6} & 28.8 & \textbf{21.7} & \textbf{37.0} & 3.25\\
\midrule
\multicolumn{6}{@{}l}{\textit{(b) Qwen3-30B-A3B (MoE), step 800}}\\
GRPO & 93.8 & 40.4 & 46.7 & 60.3 & 1.19\\
\FR & \textbf{95.6} & \textbf{44.9} & \textbf{60.8} & \textbf{67.1} & 1.94\\
\bottomrule
\end{tabular}
\end{table}

Both methods are our own runs with 128 rollouts per update: 128 prompts sampled once for \FR, and 32 prompts sampled four times for GRPO.  On Qwen2.5-7B-Instruct, GRPO stops calling the tool within roughly 200 steps; \FR maintains 3.25 calls per trajectory and leads the mean by 6.7 points (\cref{tab:tool-main}).  On Qwen3-30B-A3B, \FR leads GRPO by 6.8 points at the same 102.4k-rollout budget, at policy lag $\approx8$ against GRPO's lag $\approx1$, close to synchronous training.

\begin{figure}[!htb]
\centering
\includegraphics[width=0.9\textwidth]{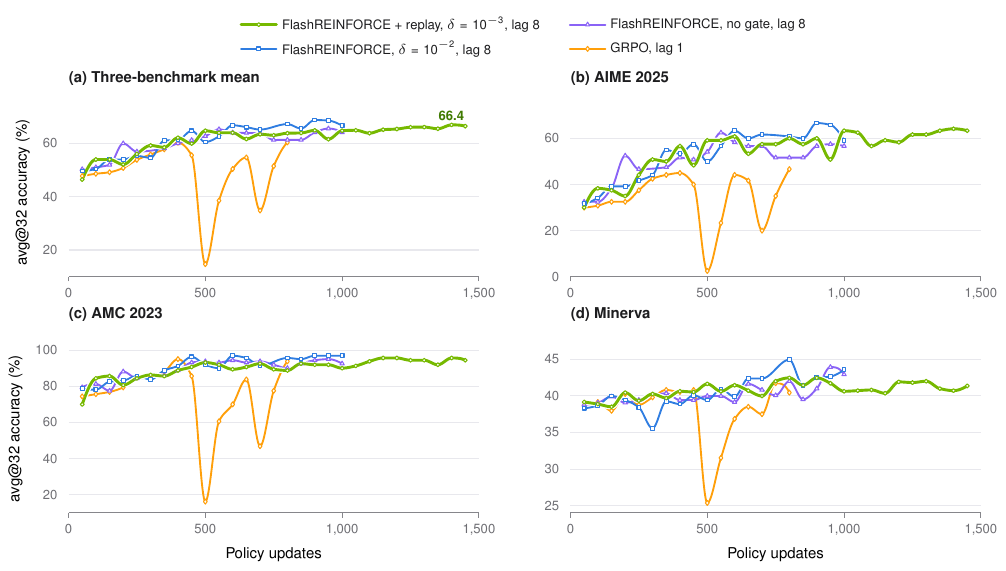}
\caption{Qwen3-30B-A3B Python-tool accuracy: \FR at lag eight and GRPO at lag one.  Markers show avg@32 evaluations, joined by shape-preserving interpolation; the mean equally weights the three benchmarks.  R3, the $\delta=10^{-3}$ configuration with routing replay, is shown through 1,450 policy updates.}
\label{fig:moe-tool-eval}
\end{figure}

To examine longer-run stability, we compare three \FR configurations at lag eight with GRPO at lag one (\cref{fig:moe-tool-eval}): the $\delta=10^{-2}$ configuration of \cref{tab:tool-main}, one without the gate, and R3, which uses $\delta=10^{-3}$ and routing replay (the learner reuses the expert routing recorded at rollout time).  R3 maintains a three-benchmark mean of 63.7--66.8 over policy updates 1,000--1,450, ending at 66.4, while GRPO suffers two sharp drops before recovering to 60.3 at step 800.  At the shared step-1,000 checkpoint, all three \FR configurations sustain tool use, with three-benchmark means of 64.0--66.5 (\cref{tab:moe-tool-aux}).

\subsection{Controlled Comparison with a Group Baseline}
\label{sec:group-batch-ablation}

Does giving up sibling rollouts cost accuracy?  We run a controlled comparison in which both arms share the settings in \cref{tab:group-batch-setup} and differ in how the 128 rollouts of each update are allocated and in which mean is subtracted.  The batch-mean arm is the \FR run of \cref{fig:r1-trust-stability}.  The group-mean arm subtracts the within-prompt mean as in Dr.\ GRPO \citep{liu2025drgrpo}, without standard-deviation normalization.  It receives the same token IS, sequence trust, and sample-mean loss, so it is a strengthened group baseline.

\begin{table}[!htb]
\caption{Setup of the controlled comparison.  The batch-mean arm is \FR.  Only the first two rows differ.}
\label{tab:group-batch-setup}
\centering
\FRtable
\begin{tabularx}{0.72\linewidth}{@{}l>{\raggedright\arraybackslash}X>{\raggedright\arraybackslash}X@{}}
\toprule
& \textbf{Batch mean} & \textbf{Group mean}\\
\midrule
Rollouts & 128 prompts $\times$ 1 & 32 prompts $\times$ 4\\
Advantage & $R_i$ minus batch mean & $R_i$ minus prompt mean\\
\midrule
Model & \multicolumn{2}{@{}l@{}}{DeepSeek-R1-Distill-Qwen-1.5B}\\
Data & \multicolumn{2}{@{}l@{}}{The same 1,460 MATH problems}\\
Loss & \multicolumn{2}{@{}l@{}}{Token IS, gate ($\delta=5\times10^{-3}$), sample mean}\\
Training & \multicolumn{2}{@{}l@{}}{Asynchronous, lag $\approx4$; AdamW, peak lr $10^{-6}$}\\
\bottomrule
\end{tabularx}
\end{table}

\begin{figure}[!htb]
\centering
\includegraphics[width=0.9\textwidth]{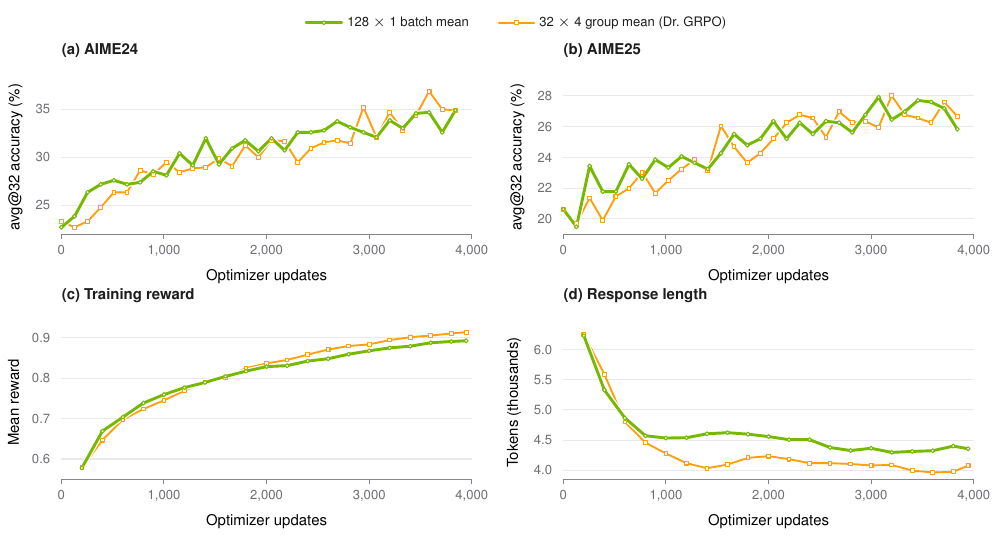}
\caption{Batch-mean and group-mean training under the \FR loss, with records through update 3,945 and AIME evaluations through update 3,840.  (a,b) avg@32 accuracy at checkpoints; (c,d) consecutive 200-update means and the final 145-update window, joined by lines.}
\label{fig:group-batch}
\end{figure}

The two arms follow similar learning curves for approximately 4,000 updates (\cref{fig:group-batch}).  Over all 30 paired evaluations, the batch-mean arm is ahead by 0.43 points in mean AIME24/25 accuracy; over the seven evaluations after update 3,000, the group-mean arm is ahead by 0.22 points (30.6 versus 30.4).  Both gaps are smaller than the typical change between consecutive checkpoints of either arm, about 0.8 points.  Single-rollout batch centering thus retains the accuracy of group centering without sibling rollouts, consistent with the fixed-budget tradeoff between a prompt-local baseline and prompt coverage \citep{hu2026groupbatch}.  The \FR loss also keeps group-based training stable under asynchrony.

\begin{table}[!htb]
\caption{Five-benchmark avg@32 accuracy (\%) on Qwen2.5-Math-1.5B.  The baselines are the published results of \citet{wu2026crf}, trained synchronously with 512k rollouts (512 per update).  \FR is trained asynchronously with 256k rollouts (128 per update), half the baseline budget.  $G$: rollouts per prompt; bold: best score.}
\label{tab:math-main}
\centering
\FRtable
\fittable[\linewidth]{%
\begin{tabular}{@{}llrrrcccccc@{}}
\toprule
\textbf{Method} & \textbf{Training} & $G$ & \textbf{Step} & \textbf{Rollouts} & \textbf{MATH-500} & \textbf{AMC23} & \textbf{Minerva} & \textbf{AIME25} & \textbf{OlympiadBench} & \textbf{Mean}\\
\midrule
Base model & -- & -- & 0 & 0 & 35.1 & 30.9 & 6.7 & 5.4 & 19.4 & 19.5\\
C-RF + NTF & Synchronous & 1 & 1,000 & 512k & 69.6 & 55.5 & 18.4 & 9.4 & 23.9 & 35.3\\
GRPO & Synchronous & 16 & 1,000 & 512k & \textbf{71.0} & \textbf{57.1} & 18.2 & 10.0 & 24.9 & 36.3\\
\textbf{\FR} & Asynchronous & 1 & 2,000 & 256k & 69.2 & 52.5 & \textbf{22.8} & \textbf{10.8} & \textbf{34.5} & \textbf{38.0}\\
\bottomrule
\end{tabular}}
\end{table}

\subsection{Published Synchronous Baselines}
\label{sec:math-main}
\label{sec:alfworld-main}

Single-rollout training targets the accuracy of group-relative training at lower cost.  The published GRPO and C-RF + NTF results come from synchronous training; \FR is trained asynchronously with half the rollouts, so the setting favors the baselines.  \FR still attains the highest five-benchmark mean on Qwen2.5-Math-1.5B: 38.0, against 36.3 for GRPO and 35.3 for C-RF + NTF (\cref{tab:math-main}).  It leads on Minerva, AIME25, and OlympiadBench, by 4.6, 0.8, and 9.6 points over GRPO.  Its mean is already 35.9 at step 1,000, with a quarter of the baseline rollouts, and stays between 37.6 and 38.0 from 256k to the full 512k rollouts (\cref{tab:long-horizon-main,tab:gate-threshold}), above both baselines under either gate threshold.  Asynchronous single-rollout training thus reaches synchronous accuracy with half the rollouts.

\begin{table}[!htb]
\caption{ALFWorld success rate (\%) with Qwen2.5-7B-Instruct.  Published rows are from \citet{wu2026crf}, trained synchronously; \FR is trained asynchronously and evaluated at step 200 after 12.8k training trajectories.}
\label{tab:alfworld-main}
\centering
\FRtable
\begin{tabular}{@{}lcrr@{}}
\toprule
\textbf{Method} & \textbf{Source} & \textbf{Seen} & \textbf{Unseen}\\
\midrule
GRPO, $G=8$ & Published & 78.6 & 76.8\\
C-RF + NTF, $G=1$ & Published & 90.5 & 86.3\\
\textbf{\FR, $G=1$} & Ours & \textbf{98.3} & \textbf{96.5}\\
\bottomrule
\end{tabular}
\end{table}

On ALFWorld, \FR improves seen/unseen success by 7.8/10.2 points over the strongest published baseline (\cref{tab:alfworld-main}).  C-RF + NTF also uses $G=1$, so prompt diversity alone does not explain the gain.

\subsection{Ablations}
\label{sec:ablations}

Each ablation changes one design choice of \cref{fig:choices} to its common alternative and keeps the others; every such change degrades or collapses training.

\paragraph{Batch centering.}
\label{sec:baseline-ablation}
Dropping the negative-advantage terms of reward-zero trajectories leaves an update that uses only successful trajectories, as REINFORCE with $0/1$ rewards and no baseline would, up to the batch-wide scale $1-\bar R$.  On Qwen2.5-7B-Instruct tool use, batch centering reaches 36.2 mean accuracy with 3.30 tool calls per trajectory; the positive-only variant reaches 9.8 and makes no tool calls (\cref{tab:baseline-ablation}, panel (a)).

\begin{table}[!htb]
\caption{Ablations on Qwen2.5-7B-Instruct tool use.  (a) Learning signal at step 700.  (b) Loss reduction, continued from the same step-900 checkpoint.  Trunc.: share of responses at the token limit.}
\label{tab:baseline-ablation}
\centering
\FRtable
\textbf{(a) Batch-centering ablation}\par\smallskip
\begin{tabular}{@{}lrrrrr@{}}
\toprule
\textbf{Learning signal} & \textbf{AMC23} & \textbf{Minerva} & \textbf{AIME25} & \textbf{Mean} & \textbf{Calls}\\
\midrule
\textbf{Batch-centered} & \textbf{57.5} & \textbf{30.2} & \textbf{20.8} & \textbf{36.2} & \textbf{3.30}\\
Positive-only & 12.5 & 16.9 & 0.0 & 9.8 & 0.00\\
\bottomrule
\end{tabular}

\medskip
\textbf{(b) Loss reduction after step 900}\par\smallskip
\begin{tabular}{@{}lrrrrr@{}}
\toprule
\textbf{State} & \textbf{Updates} & \textbf{Reward} & \textbf{Length} & \textbf{Calls} & \textbf{Trunc.}\\
\midrule
Step-900 reference & 0 & 0.430 & 2,573 & 3.87 & 18.8\\
\textbf{Sample mean} & $\sim150$ & \textbf{0.477} & \textbf{2,518} & \textbf{4.01} & \textbf{17.2}\\
Token mean & $\sim150$ & 0.403 & 3,947 & 2.18 & 45.3\\
\bottomrule
\end{tabular}
\end{table}

\paragraph{Loss reduction.}
\label{sec:reduction-ablation}
We continue both reductions from the same step-900 checkpoint for roughly 150 steps.  Sample mean maintains reward and tool use, whereas token mean enters a long-response collapse: responses grow from 2,573 to 3,947 tokens and truncation surges from 18.8\% to 45.3\%, with fewer tool calls (\cref{tab:baseline-ablation}, panel (b)).  The reward has barely moved (0.430 to 0.403), but with nearly half of the responses at the token limit the run is no longer trainable.

\begin{table}[!htb]
\caption{Sequence-level versus token-level admission on Qwen2.5-Math-1.5B with the same drift proxy and threshold ($\delta=10^{-3}$).  Arrows show peak to last value.}
\label{tab:trust-granularity}
\centering
\FRtable
\begin{tabular}{@{}lrrrl@{}}
\toprule
\textbf{Admission} & \textbf{Steps} & \textbf{AMC23} & \textbf{Train reward} & \textbf{Outcome}\\
\midrule
\textbf{Sequence-level} & 2,160 & 55.6 $\rightarrow$ 53.1 & 0.23 & \textbf{Stable}\\
Token-level & 1,237 & 51.9 $\rightarrow$ 26.3 & 0.18 $\rightarrow$ 0.01 & Collapsed\\
\bottomrule
\end{tabular}
\end{table}

\paragraph{Admission granularity.}
\label{sec:trust-granularity-ablation}
On Qwen2.5-Math-1.5B we apply the same drift proxy and threshold, $\delta=1\times10^{-3}$, either to whole trajectories or to individual tokens (keeping token $t$ iff $d_{i,t}\leq\delta$).  Sequence-level admission stays stable through 2,160 steps, with AMC23 at 53.1 and training reward 0.23.  Token-level masking trains to 51.9 on AMC23 and then collapses by step 1,237: AMC23 falls to 26.3, training reward from 0.18 to 0.01, and entropy rises from 0.37 to 5.6 (\cref{tab:trust-granularity}).  The reason is that of \cref{sec:local-surrogate}: a masked token still shaped the contexts of the tokens that remain (\cref{app:granularity-theory}).  DPPO likewise reports that IS alone can fail as training continues \citep[][Sec.~5.1 and Figure~3]{qi2026dppo}.

\begin{table}[!htb]
\caption{Sequence trust on DeepSeek-R1-Distill-Qwen-1.5B at policy lag 4, with token IS in both runs.  Accuracy is the AIME24/25 avg@32 mean; KL is the sampled-action proxy, averaged over updates 4,801--5,000 (first two rows) and 5,801--6,000 (last two).}
\label{tab:r1-trust-summary}
\centering
\FRtable
\begin{tabular}{@{}lrrr@{}}
\toprule
\textbf{Setting} & \textbf{Step} & \textbf{Acc.\ (\%)} & \textbf{KL ($\times10^{-4}$)}\\
\midrule
With sequence trust & 4,992 & \textbf{32.2} & \textbf{6.73}\\
Without trust & 4,992 & 30.6 & 7.38\\
\addlinespace[3pt]
With sequence trust & 5,896 & \textbf{33.7} & \textbf{8.96}\\
Without trust & 5,888 & 30.6 & 9.45\\
\bottomrule
\end{tabular}
\end{table}

\paragraph{With and without the gate.}
\label{sec:sequence-trust-region-ablation}
On DeepSeek-R1-Distill-Qwen-1.5B at lag four, sequence trust reaches 33.7 AIME24/25 mean accuracy at step 5,896, versus 30.6 without trust at higher measured drift, and leads at all seven evaluations after update 5,000 (\cref{tab:r1-trust-summary,tab:r1-aime-scores}).  On ALFWorld, jointly removing IS and sequence trust drops step-40 seen/unseen success from 81.0/77.6 to 35.0/27.6.  Both $\delta=1\times10^{-3}$ and $\delta=3\times10^{-3}$ keep Qwen2.5-Math training stable for 4,000 steps (\cref{tab:gate-threshold}); the latter rejects 0.13\% of trajectories.

\paragraph{Fresh one-batch updates.}
\label{sec:one-batch-ablation}
We compare collecting 128 completed trajectories per update with collecting 512 under one behavior policy and processing four sequential minibatches of 128.  Both schedules use each trajectory once with the same IS correction and gate.  In the sequential schedule, the last three minibatches meet a learner that has advanced one to three updates beyond their behavior policy.  At equal data and optimization budget on Qwen2.5-Math-1.5B (102.4k trajectories, 800 optimizer steps), fresh updates reach 28.45 mean accuracy on AMC23, Minerva, and AIME25, versus 26.80 for sequential minibatches.

\section{Related Work}
\label{sec:related}

\paragraph{Baselines for language-model RL.}
REINFORCE uses score-function updates, with action-independent baselines reducing variance \citep{williams1992reinforce,sutton2000policygradient,greensmith2004variance}.  RLOO and GRPO use sibling-rollout baselines \citep{ahmadian2024rloo,shao2024deepseekmath}, and Dr.\ GRPO removes their length and standard-deviation normalization \citep{liu2025drgrpo}.  REINFORCE++, DAPO, and CISPO refine normalization, clipping, and sampling \citep{hu2025reinforcepp,yu2025dapo,minimax2025m1}, and \citet{liu2025tricks} recommend group-level means with token-level loss.  C-RF + NTF protects high-probability tokens from negative feedback \citep{wu2026crf}.  SAO uses a learned value baseline for asynchronous single-rollout training \citep{hou2026sao}, while SPO maintains per-prompt statistics \citep{xu2025spo}.

\paragraph{Trust regions and divergence control.}
Natural policy gradient and TRPO motivate local divergence control, while PPO uses sampled-ratio clipping \citep{amari1998natural,kakade2001natural,schulman2015trpo,schulman2017ppo}.  DPPO studies divergence-based proximal updates \citep{qi2026dppo}; GSPO uses sequence ratios and cumulative-token policy optimization studies prefix correction \citep{zheng2025gspo,zhang2026ctpo}.  Seq-MIS and Trust Region Masking mask whole sequences by a sequence ratio and by the largest token KL, respectively \citep{liu2025seqmis,li2026trustregionmasking}; \FR gates on a trajectory-mean KL proxy.

\paragraph{Asynchronous and off-policy learning.}
IMPALA established importance-corrected actor--learner training \citep{espeholt2018impala}.  Asynchronous RLHF, AReaL, and LlamaRL study throughput, staleness, and synchronization for language models \citep{noukhovitch2025asyncrlhf,fu2025areal,wu2025llamarl}.  GiGPO studies long-horizon credit assignment \citep{feng2025gigpo}.  \citet{zhong2026tim,zhang2025tbik,zhang2026prls1} study training--inference mismatch.

\section{Conclusion}

\FR shows that critic-free single-rollout RL scales when three choices are made together: batch centering on fresh batches, sequence-level trust, and sample-mean weighting, under which a batch-mean baseline is stable.  It trains stably through 6,000 asynchronous updates at lag $\approx4$ and through 1,450 updates on a 30B MoE at lag eight, matches group-mean centering in a controlled comparison, and reaches the accuracy of synchronous published baselines with half the rollouts.  Agent training thus has a simple option that needs neither sibling rollouts nor a value model.  Future work includes drift statistics that also track unobserved actions (\cref{app:granularity-theory}) and longer-horizon environments such as OSWorld \citep{xie2024osworld}.

\section*{Limitations}

\FR is designed as a compute-efficient recipe, not as the strongest method in every configuration.  It uses one rollout per prompt and no critic, and assigns one trajectory-level advantage to every token of a trajectory.  Methods that spend more compute can do better in some settings; for example, a learned value model \citep{hou2026sao} can refine credit assignment across the turns of a multi-turn task.  Our results establish that the critic-free single-rollout setting trains stably at scale and keeps the accuracy of a group baseline, and they make no claim of the best attainable accuracy for every model, task, or budget.

\bibliography{reference}
\bibliographystyle{plainnat}

\clearpage
\appendix
\raggedbottom
\makeatletter
\setlength{\@fptop}{0pt}
\setlength{\@fpsep}{12pt plus 2pt}
\setlength{\@fpbot}{0pt plus 1fil}
\makeatother
\section{Experimental Details and Additional Results}
\label{app:experimental-details}

\subsection{Long-CoT reasoning: DeepSeek-R1-Distill-Qwen-1.5B}
\label{app:r1-details}

We train DeepSeek-R1-Distill-Qwen-1.5B \citep{deepseekai2025r1} on DPPO's 1,460-problem MATH subset \citep{qi2026dppo}, at policy lag $\approx4$ with 128 trajectories per update, and report mean AIME24/25 avg@32 accuracy.  \Cref{tab:r1-config} lists the hyperparameters, and \cref{tab:r1-aime-scores} the AIME24 and AIME25 scores behind \cref{fig:r1-trust-stability,tab:r1-trust-summary}.

\begin{table}[!htb]
\caption{Hyperparameters for the DeepSeek-R1-Distill-Qwen-1.5B trust comparison.}
\label{tab:r1-config}
\centering
\footnotesize
\FRtableformat
\begin{tabularx}{\linewidth}{@{}
  >{\raggedright\arraybackslash}p{0.23\linewidth}
  >{\raggedright\arraybackslash}X@{}}
\toprule
\multicolumn{2}{@{}l}{\textbf{Data and sampling}}\\
\cmidrule{1-2}
Training data & \texttt{sail/Sanity-Test-R1D-1.5B}: 1,460 MATH problems with initial solve rates of 20--80\% \citep{qi2026dppo}\\
Token budget & Prompt: 1,024; response: 8,192; total: 9,216 tokens\\
Rollout sampling & Temperature $1.0$; top-$p=1.0$\\
Evaluation & AIME24 and AIME25; 30 problems each; 32 samples per problem; every 128 updates from step 0\\
Evaluation sampling & Temperature $0.6$; top-$p=0.95$; 8,192 generated tokens\\
\addlinespace[4pt]
\multicolumn{2}{@{}l}{\textbf{Update and optimization}}\\
\cmidrule{1-2}
Online update & 128 prompts, one rollout each; batch-mean baseline; sample-mean loss; token IS\\
Optimizer & AdamW; peak lr $10^{-6}$; weight decay $0.1$; gradient clip $1.0$\\
LR schedule & Cosine decay with extended horizon; 3\% initial warmup; minimum lr $10^{-7}$\\
\addlinespace[4pt]
\multicolumn{2}{@{}l}{\textbf{Asynchronous collection and sequence trust}}\\
\cmidrule{1-2}
Collection & 512-prompt generation requests; 128-trajectory learner batches; queue depth 8 (policy lag $\approx4$)\\
Sequence threshold & $\delta=5\times10^{-3}$\\
\bottomrule
\end{tabularx}
\end{table}

\begin{table}[!htb]
\caption{AIME avg@32 accuracy (\%).  (a) Trust comparison through 5,000 updates.  (b) Continued training within 6,000 cumulative updates; approximate steps align nearby evaluations across arms.  Mean equally weights AIME24 and AIME25.  Boldface marks the with-trust run's AIME24 peak.}
\label{tab:r1-aime-scores}
\centering\footnotesize\FRtableformat
\textbf{(a) With versus without sequence trust}\par\smallskip
\begin{tabular}{@{}rrrrrrr@{}}
\toprule
& \multicolumn{3}{c}{\textbf{With sequence trust}} & \multicolumn{3}{c}{\textbf{Without trust (IS only)}}\\
\cmidrule(lr){2-4}\cmidrule(l){5-7}
\textbf{Step} & \textbf{AIME24} & \textbf{AIME25} & \textbf{Mean} & \textbf{AIME24} & \textbf{AIME25} & \textbf{Mean}\\
\midrule
0 & 22.71 & 20.63 & 21.67 & 22.71 & 20.42 & 21.56\\
1,024 & 28.13 & 23.33 & 25.73 & 27.60 & 24.06 & 25.83\\
2,048 & 31.98 & 26.35 & 29.17 & 28.85 & 25.63 & 27.24\\
3,072 & 32.08 & 27.92 & 30.00 & 33.13 & 26.46 & 29.79\\
3,968 & 35.42 & 26.25 & 30.83 & 34.58 & 26.77 & 30.68\\
4,992 & 36.15 & 28.33 & 32.24 & 33.85 & 27.40 & 30.63\\
\bottomrule
\end{tabular}\par\medskip
\textbf{(b) Continued training beyond 5,000 updates}\par\smallskip
\begin{tabular}{@{}rrrrrrr@{}}
\toprule
& \multicolumn{3}{c}{\textbf{With sequence trust}} & \multicolumn{3}{c}{\textbf{Without trust (IS only)}}\\
\cmidrule(lr){2-4}\cmidrule(l){5-7}
\textbf{Step ($\approx$)} & \textbf{AIME24} & \textbf{AIME25} & \textbf{Mean} & \textbf{AIME24} & \textbf{AIME25} & \textbf{Mean}\\
\midrule
5,100 & 34.79 & 27.92 & 31.35 & 31.98 & 26.25 & 29.11\\
5,300 & 35.10 & 29.48 & 32.29 & 33.96 & 28.33 & 31.15\\
5,400 & 36.56 & 27.08 & 31.82 & 32.81 & 27.19 & 30.00\\
5,500 & 35.00 & 27.50 & 31.25 & 33.96 & 26.98 & 30.47\\
5,600 & 35.94 & 28.23 & 32.08 & 32.60 & 27.81 & 30.21\\
5,800 & 35.31 & 28.65 & 31.98 & 31.67 & 28.02 & 29.84\\
5,900 & \textbf{38.85} & 28.54 & 33.70 & 33.33 & 27.81 & 30.57\\
\bottomrule
\end{tabular}
\end{table}

\subsection{Python tool use}
\label{app:tool-details}

We train Qwen2.5-7B-Instruct and the 30B MoE Qwen3-30B-A3B with a Python tool \citep{qwen2025qwen25,yang2025qwen3}.  Each update uses 128 rollouts: 128 prompts sampled once for \FR, or 32 prompts sampled four times for GRPO.  We evaluate committed answers on AMC23, Minerva, and AIME25.  \Cref{tab:tool-config,tab:moe-tool-config} list the settings.  All three \FR configurations in \cref{tab:moe-tool-aux} sustain tool use at lag eight; the main comparison in \cref{tab:tool-main} uses the first row's setting.

\begin{table}[!htb]
\caption{Configuration of the Qwen2.5-7B-Instruct Python-tool experiment.}
\label{tab:tool-config}
\centering\footnotesize\FRtableformat
\begin{tabularx}{\linewidth}{@{}>{\raggedright\arraybackslash}p{0.23\linewidth}>{\raggedright\arraybackslash}X@{}}
\toprule
\textbf{Setting} & \textbf{Value}\\
\midrule
Training data & DAPO-style math prompts\\
Online batch & 128 prompts $\times$ one rollout\\
Interaction budget & 10 tool turns; 8,192 tokens at comparison\\
Generation budget & 6,144 tokens across turns\\
Reward & Exact committed answer\\
Optimization & AdamW; cosine lr $10^{-6}$\\
Sequence threshold & $\delta=1\times10^{-3}$\\
Evaluation & AMC23, Minerva, AIME25; avg@32; temperature $0.7$; top-$p=0.7$\\
\bottomrule
\end{tabularx}
\end{table}

\begin{table}[!htb]
\caption{Configuration of the Qwen3-30B-A3B Python-tool experiments.}
\label{tab:moe-tool-config}
\centering\footnotesize\FRtableformat
\begin{tabularx}{\linewidth}{@{}>{\raggedright\arraybackslash}p{0.23\linewidth}>{\raggedright\arraybackslash}X@{}}
\toprule
\textbf{Setting} & \textbf{Value}\\
\midrule
Model & Qwen3-30B-A3B; thinking disabled\\
Training data & Deduplicated 7.5k DAPO-Math prompts, formatted for tool use\\
Rollouts per update & \FR: 128 prompts $\times1$; GRPO: 32 prompts $\times4$\\
Policy lag & \FR: $\approx8$; GRPO: $\approx1$\\
Interaction budget & Up to 10 turns; 8,192 generated tokens per turn; 18,432 total tokens\\
Rollout sampling & Temperature $1.0$; top-$p=1.0$\\
Optimization & AdamW; initial lr $10^{-6}$ with cosine decay; weight decay $0.1$\\
Main comparison & $\delta=10^{-2}$; routing replay off\\
Longer stability run & $\delta=10^{-3}$; routing replay on (R3)\\
GRPO update & Group-normalized advantages; PPO clipping $0.2$; routing replay off\\
Reference KL & Disabled in all configurations\\
Evaluation & AMC23, Minerva, AIME25; avg@32; temperature $0.7$; top-$p=0.7$; every 50 updates\\
\bottomrule
\end{tabularx}
\end{table}

\begin{table}[!htb]
\caption{MoE configuration comparison after 1,000 policy updates, all at policy lag eight.  Accuracy is avg@32 (\%), with an equally weighted three-task mean; calls per trajectory and truncation rate (\%) are averaged over updates 901--1,000.  Replay denotes MoE routing replay.}
\label{tab:moe-tool-aux}
\centering\footnotesize\FRtableformat
\fittable{%
\begin{tabular}{@{}lcrrrrrr@{}}
\toprule
\textbf{Gate $\delta$} & \textbf{Replay} & \textbf{AMC23} & \textbf{Minerva} & \textbf{AIME25} & \textbf{Mean} & \textbf{Calls} & \textbf{Trunc.}\\
\midrule
$10^{-2}$ & Off & 96.9 & 43.6 & 59.2 & 66.5 & 1.76 & 7.2\\
$10^{-3}$ & On & 90.0 & 40.6 & 63.3 & 64.7 & 1.95 & 2.8\\
No gate & Off & 92.5 & 42.9 & 56.7 & 64.0 & 1.80 & 7.6\\
\bottomrule

\end{tabular}}
\end{table}

\FloatBarrier
\subsection{Mathematical reasoning: Qwen2.5-Math-1.5B}
\label{app:qwen-details}
\label{app:additional-results}
\label{app:long-horizon-dynamics}

We train Qwen2.5-Math-1.5B \citep{yang2024qwen25math} on 7.5k deduplicated DAPO-Math prompts \citep{yu2025dapo} and report avg@32 on MATH-500 \citep{hendrycks2021math}, AMC23, Minerva \citep{lewkowycz2022minerva}, AIME25, and OlympiadBench \citep{he2024olympiadbench}.  \Cref{tab:math-config} lists the settings, and \cref{tab:long-horizon-main} the checkpoints through step 2,000, the checkpoint reported in \cref{tab:math-main}.

\begin{table}[H]
\caption{Qwen2.5-Math baseline-comparison configuration.  Boldface marks the defining one-pass design choices.}
\label{tab:math-config}
\centering
\small
\FRtableformat
\begin{tabularx}{\linewidth}{@{}lXlX@{}}
\toprule
\multicolumn{4}{@{}l}{\textbf{Model and Data}}\\[-2pt]
\cmidrule(r){1-4}
Model & Qwen2.5-Math-1.5B
& Prompts & Deduplicated 7.5k DAPO-Math\\
Token budget & 2,048 prompt; 4,096 response
& Online batch & 128 prompts; \textbf{one rollout each}\\
\addlinespace[4pt]
\multicolumn{4}{@{}l}{\textbf{Optimization}}\\[-2pt]
\cmidrule(r){1-4}
Update & \textbf{One full-batch step; no minibatches or reuse}
& Optimizer & AdamW; lr $10^{-6}$; wd $0.1$\\
Schedule & Cosine; gradient clip $1.0$
& Sequence threshold & $\delta=3\times10^{-3}$\\
\addlinespace[4pt]
\multicolumn{4}{@{}l}{\textbf{Collection and Sampling}}\\[-2pt]
\cmidrule(r){1-4}
Collection & Asynchronous; one rollout per prompt
& Policy lag & Set by asynchronous completion order\\
Rollout & Temperature $1.0$; top-$p=1.0$
& Evaluation & Temperature $0.7$; top-$p=0.7$\\
\bottomrule
\end{tabularx}
\end{table}

\begin{table}[H]
\caption{Qwen2.5-Math-1.5B checkpoints through step 2,000 with $\delta=3\times10^{-3}$: avg@32 accuracy (\%) and the five-benchmark mean.  Boldface marks the checkpoint reported in \cref{tab:math-main}.}
\label{tab:long-horizon-main}
\centering
\footnotesize
\FRtableformat
\begin{tabular}{@{}rrrrrrrr@{}}
\toprule
\textbf{Step} & \textbf{Rollouts} & \textbf{MATH-500} & \textbf{AMC23} & \textbf{Minerva} & \textbf{AIME25} & \textbf{OlympiadBench} & \textbf{Mean}\\
\midrule
1,000 & 128k & 66.4 & 50.5 & 21.3 & 9.0 & 32.5 & 35.94\\
1,500 & 192k & 68.2 & 51.6 & 22.8 & 10.2 & 33.6 & 37.29\\
\textbf{2,000} & \textbf{256k} & \textbf{69.2} & \textbf{52.5} & \textbf{22.8} & \textbf{10.8} & \textbf{34.5} & \textbf{37.97}\\
\bottomrule
\end{tabular}
\end{table}

\paragraph{Gate threshold.}
\label{app:gate-threshold}
We examine sensitivity to the gate threshold by comparing the $\delta=1\times10^{-3}$ and $\delta=3\times10^{-3}$ configurations on Qwen2.5-Math-1.5B.  Both runs use one rollout per prompt, 128 trajectories per update, and sample-mean reduction.  We evaluate the same four checkpoints on all five reasoning benchmarks with 32 samples per prompt, temperature $0.7$, and top-$p=0.7$.

\begin{table}[H]
\caption{Gate-threshold configurations on Qwen2.5-Math-1.5B.  Entries are five-benchmark mean avg@32 accuracy (\%); the four checkpoints use 128k, 256k, 384k, and 512k training rollouts.  Boldface marks the higher mean at each checkpoint.}
\label{tab:gate-threshold}
\centering
\footnotesize
\FRtableformat
\begin{tabular}{@{}lrrrr@{}}
\toprule
$\delta$ & \textbf{Step 1,000} & \textbf{Step 2,000} & \textbf{Step 3,000} & \textbf{Step 4,000}\\
\midrule
$1\times10^{-3}$ & \textbf{36.04} & 36.89 & 36.93 & 37.09\\
$3\times10^{-3}$ & 35.94 & \textbf{37.97} & \textbf{37.75} & \textbf{37.58}\\
\bottomrule
\end{tabular}
\end{table}

Both configurations remain stable through 4,000 updates.  The looser threshold yields higher means at steps 2,000--4,000 in these runs, consistent with using the gate as a safeguard against large proxy-drift outliers.

\subsection{ALFWorld: Qwen2.5-7B-Instruct}
\label{app:alfworld-details}

We train Qwen2.5-7B-Instruct on 1,024 ALFWorld games with 64 trajectories per update and evaluate 140 seen and 134 unseen games, matching the published protocol \citep{shridhar2021alfworld,wu2026crf}.  \Cref{tab:alfworld-config} lists the settings.

\begin{table}[!htb]
\caption{Configuration of the Qwen2.5-7B-Instruct ALFWorld experiment.}
\label{tab:alfworld-config}
\centering\footnotesize\FRtableformat
\begin{tabularx}{\linewidth}{@{}>{\raggedright\arraybackslash}p{0.23\linewidth}>{\raggedright\arraybackslash}X@{}}
\toprule
\textbf{Setting} & \textbf{Value}\\
\midrule
Training data & 1,024 training games\\
Online batch & 64 games $\times$ one rollout\\
Interaction budget & 50 environment actions; 16,384 total tokens\\
Generation budget & 8,192 tokens across turns\\
Reward & Terminal environment success\\
Optimization & AdamW; constant lr $10^{-6}$\\
Sequence threshold & $\delta=1\times10^{-3}$\\
Evaluation & 140 seen + 134 unseen games; temperature $1.0$; top-$p=0.7$\\
\bottomrule
\end{tabularx}
\end{table}

\section{Token-Level versus Sequence-Level Trust Regions}
\label{app:granularity-theory}

The Qwen2.5-Math comparison in \cref{tab:trust-granularity} shows token-local collapse while sequence-level admission remains stable.  This appendix gives two reasons for making one decision per trajectory: local-surrogate error accumulates policy movement across the sequence, and removing one token's score term leaves later stored histories unchanged.  We use a fixed policy-token horizon $T$ with absorbing padding after termination, and a shared prompt distribution and environment for $\pi$ and $\mu$; $T_i$ denotes an observed trajectory's length.

\paragraph{Local surrogate.}
Write $J(\pi)=\mathbb E_{\tau\sim\pi}[R(\tau)]$ for the return, $d_\mu^t$ for the distribution of the history $h_t$ under a behavior policy $\mu$, and $Q_t^\mu(h,a)=\mathbb E_\mu[R\mid h_t=h,a_t=a]$ for the behavior continuation value.  Define $V_t^\mu(h)=\mathbb E_\mu[R\mid h_t=h]$ and the behavior advantage $A_t^\mu(h,a)=Q_t^\mu(h,a)-V_t^\mu(h)$, with $|A_t^\mu|\leq A_{\max}$.  Holding these behavior quantities fixed, the local surrogate of \cref{sec:local-surrogate} is
\begin{equation}
 \mathcal L_\mu(\pi)
 =J(\mu)+\sum_{t}\mathbb E_{h\sim d_\mu^t}\,
 \mathbb E_{a\sim\pi(\cdot\mid h)}\big[A_t^\mu(h,a)\big].
 \label{eq:local-surrogate}
\end{equation}
The additive $J(\mu)$ makes $\mathcal L_\mu(\mu)=J(\mu)$.  Histories and continuation values remain behavioral.  The per-position movements are $\varepsilon_t=\sup_h\TV(\pi(\cdot\mid h),\mu(\cdot\mid h))$ and $\kappa_t=\sup_h\KL(\mu(\cdot\mid h)\,\|\,\pi(\cdot\mid h))$, with $\bar\kappa=T^{-1}\sum_{t=1}^{T}\kappa_t$.  Let $f_t(h)=\mathbb E_{a\sim\pi(\cdot\mid h)}[A_t^\mu(h,a)]$.  Then \cref{eq:local-surrogate} becomes
\begin{equation}
 \mathcal L_\mu(\pi)
 =J(\mu)+\sum_{t=1}^{T}\mathbb E_{h\sim d_\mu^t}\big[f_t(h)\big],
 \label{eq:app-local-surrogate}
\end{equation}

\begin{prop}[Exact local gradient]
\label[prop]{prop:local-gradient}
For a differentiable policy satisfying $\pi_\theta(\cdot\mid h)\ll\mu(\cdot\mid h)$ at every behavior-reachable history, \cref{eq:local-gradient} holds, and at $\pi_\theta=\mu$ the expression equals $\nabla_\theta J(\pi_\theta)$.
\end{prop}

\begin{proof}
The baseline term vanishes because $V_t^\mu(h)$ is action-independent and $\sum_a\nabla_\theta\pi_\theta(a\mid h)=0$.  Differentiate the remaining inner expectation of \cref{eq:local-surrogate} with $d_\mu^t$ and $Q_t^\mu$ held fixed: $\nabla_\theta\sum_a\pi_\theta(a\mid h)Q_t^\mu(h,a)=\mathbb E_{a\sim\mu}[\rho_t Q_t^\mu(h,a)\nabla_\theta\log\pi_\theta(a\mid h)]$, using $\nabla\pi_\theta=\pi_\theta\nabla\log\pi_\theta$ and absolute continuity for the change of measure.  Replacing $Q_t^\mu(h_t,a_t)$ by the sampled $R$, whose conditional expectation it is, gives \cref{eq:local-gradient}.  At $\pi_\theta=\mu$ every ratio is one and the expression is the REINFORCE gradient of $J$.
\end{proof}

The following bound follows the classical performance-difference and coupling argument \citep{kakade2002cpi,schulman2015trpo}, in the finite-horizon form reviewed by \citet[][Sec.~3.1]{li2026trustregionmasking}, and retains per-position divergences.

\begin{prop}[Surrogate error is governed by accumulated drift]
\label[prop]{prop:accumulated-drift}
\begin{equation}
 |J(\pi)-\mathcal L_\mu(\pi)|
 \leq 4A_{\max}\sum_{t=1}^{T}\varepsilon_t\sum_{u<t}\varepsilon_u
 \leq 2A_{\max}\Big(\sum_{t=1}^{T}\varepsilon_t\Big)^{2}
 \leq A_{\max}T^{2}\bar\kappa.
 \label{eq:accumulated-drift}
\end{equation}
\end{prop}

\begin{proof}
The finite-horizon performance-difference identity gives $J(\pi)-J(\mu)=\sum_t\mathbb E_{h\sim d_\pi^t}[f_t(h)]$, so
\begin{equation}
 J(\pi)-\mathcal L_\mu(\pi)
 =\sum_{t=1}^{T}\Big(\mathbb E_{d_\pi^t}[f_t]-\mathbb E_{d_\mu^t}[f_t]\Big).
 \label{eq:app-pdl}
\end{equation}
Since $\mathbb E_{a\sim\mu}[A_t^\mu(h,a)]=0$, the function $f_t$ satisfies $|f_t(h)|\leq 2A_{\max}\TV(\pi(\cdot\mid h),\mu(\cdot\mid h))\leq 2A_{\max}\varepsilon_t$.  Coupling the two processes with shared environment randomness and a maximal coupling of the two action distributions at every history they share, the histories first disagree at step $u$ with probability at most $\varepsilon_u$, so $\TV(d_\pi^t,d_\mu^t)\leq\sum_{u<t}\varepsilon_u$.  Each term of \cref{eq:app-pdl} is bounded by $2\|f_t\|_\infty\TV(d_\pi^t,d_\mu^t)$, which gives the first inequality.  The second follows from $\sum_t\varepsilon_t\sum_{u<t}\varepsilon_u\leq\frac12(\sum_t\varepsilon_t)^2$.  For the third, Pinsker gives $\varepsilon_t\leq\sqrt{\kappa_t/2}$ and Cauchy--Schwarz gives $\sum_t\sqrt{\kappa_t/2}\leq\sqrt{T\sum_t\kappa_t/2}=T\sqrt{\bar\kappa/2}$.
\end{proof}

Under $\varepsilon_t\leq\varepsilon$, the first bound gives $2A_{\max}T(T-1)\varepsilon^2$, recovering that classical coupling form.  The per-position TV bound accumulates movement across all earlier positions, and for a fixed horizon the final KL bound depends on the sum $\sum_t\kappa_t=T\bar\kappa$.  This motivates monitoring policy movement over complete trajectories; \cref{prop:expected-drift} bounds the same error by expected drift over behavior histories, without the supremum.  The computable statistic and the admission rule below are practical design choices guided by this accumulated-drift view.

\paragraph{Computable sequence trust approximation.}
The worst-case quantity $\bar\kappa$ is not directly available to the learner because it requires full behavior distributions and a supremum over behavior-reachable histories.  For the observed length $T_i$ of trajectory $i$, define the corresponding worst-case average $\bar\kappa_i=T_i^{-1}\sum_{t=1}^{T_i}\sup_h\KL(\mu_i(\cdot\mid h)\,\|\,\pi_\theta(\cdot\mid h))$ and the realized-history statistic
\begin{equation}
 D_i^{\mathrm{full}}
 =\frac1{T_i}\sum_{t=1}^{T_i}
 \KL\!\left(\mu_i(\cdot\mid h_{i,t})\,\|\,\pi_\theta(\cdot\mid h_{i,t})\right).
 \label{eq:app-full-sequence-drift}
\end{equation}
At every stored history, $d_{i,t}$ is the KL after projecting the vocabulary onto the sampled action versus all remaining actions.  The data-processing inequality and the definition of the history-wise supremum therefore give
\begin{equation}
 \bar D_i
 \leq D_i^{\mathrm{full}}
 \leq \bar\kappa_i.
 \label{eq:projected-trust-relaxation}
\end{equation}
By \cref{eq:projected-trust-relaxation}, rejecting $\bar D_i>\delta$ identifies a trajectory whose mean full KL also exceeds $\delta$.  Passing the screen provides no corresponding upper bound: probability mass can shift among unobserved actions while leaving the sampled-action probability unchanged.  The proxy supplies a low-cost drift screen using stored behavior probabilities.

\paragraph{Trajectory-normalized practical surrogate.}
\label{app:loss-reductions}

Let $\ell_{i,t}$ denote the per-token loss in \cref{eq:flash-loss}, including the fixed advantage and sequence mask, and let $L_i=T_i^{-1}\sum_{t=1}^{T_i}\ell_{i,t}$ be its within-trajectory mean.  Using the same token losses, the two reductions are
\begin{equation}
 \mathcal L_{\mathrm{sample}}=\frac1B\sum_{i=1}^{B}L_i,
 \qquad
 \mathcal L_{\mathrm{token}}
 =\frac{\sum_{i=1}^{B}\sum_{t=1}^{T_i}\ell_{i,t}}{\sum_{j=1}^{B}T_j}
 =\sum_{i=1}^{B}\frac{T_i}{\sum_{j=1}^{B}T_j}L_i.
 \label{eq:loss-reductions}
\end{equation}
Sample mean assigns each trajectory weight $1/B$; token mean assigns $T_i/\sum_jT_j$, giving longer failures greater weight even at the same mean token loss.  The token-level signal is unchanged.  With batch-centered advantages, the advantage weights of sample mean sum to $\frac1B\sum_iA_i=0$, and those of token mean to
\begin{equation}
 \begin{split}
 \frac{\sum_iA_iT_i}{\sum_jT_j}
 &=\frac{1}{B\bar T}\sum_{i=1}^{B}(R_i-\bar R)(T_i-\bar T)\\
 &=\frac{\Cov(R,T)}{\bar T},
 \qquad \bar T=\frac1B\sum_{j=1}^{B}T_j,
 \end{split}
 \label{eq:reward-length-covariance}
\end{equation}
using $\sum_iA_i\bar T=0$.  With a group mean, $A_i$ sums to zero within each prompt, and the same step leaves only the within-prompt covariances, which are small when the responses to a prompt have similar lengths.

The sample-mean update replaces the local surrogate's token sum with a within-trajectory average, changing its length weighting.  \Cref{tab:baseline-ablation} compares sample-mean and token-mean reductions from the same checkpoint.

Applying the trajectory normalization to the local gradient in \cref{eq:local-gradient}, with $A_i$ and $m_i$ held fixed, gives the update direction
\begin{equation}
 \widehat g_{\mathrm{FlashREINFORCE}}(\theta)
 =\frac1B\sum_{i=1}^{B}\frac{m_iA_i}{T_i}
 \sum_{t=1}^{T_i}\rho_{i,t}(\theta)
 \nabla_\theta\log\pi_\theta(a_{i,t}\mid h_{i,t}).
 \label{eq:flash-gradient}
\end{equation}
The detached-ratio loss in \cref{eq:flash-loss} has this negative gradient, since
\begin{equation}
 \nabla_\theta\,\rho_{i,t}(\theta)
 =\rho_{i,t}(\theta)\,\nabla_\theta\log\pi_\theta(a_{i,t}\mid h_{i,t})
 =\nabla_\theta\big[\sg(\rho_{i,t})\log\pi_\theta(a_{i,t}\mid h_{i,t})\big].
 \label{eq:sg-equivalence}
\end{equation}

\paragraph{Token gating.}
A token-local rule uses $g_{i,t}=\ind[d_{i,t}\leq\delta]$ and, with the gate held fixed, replaces the complete trajectory contribution by
\begin{equation}
 G_i^{\mathrm{tok}}
 =\frac{A_i}{BT_i}\sum_{t=1}^{T_i}g_{i,t}\rho_{i,t}
 \nabla_\theta\log\pi_\theta(a_{i,t}\mid h_{i,t}).
 \label{eq:token-gated-gradient}
\end{equation}
This operation deletes selected score terms but does not alter the stored prefix $h_{i,t}$ of any later retained position.  In particular, dropping position $u$ leaves every later learner probability $\pi_\theta(a_{i,t}\mid h_{i,t})$ evaluated at the same behavior-generated history.  The remainder of a trajectory can therefore keep contributing even when its sequence mean exceeds $\delta$.  Sequence admission instead multiplies the complete token-score sum by $m_i$: every admitted trajectory satisfies $\bar D_i\leq\delta$, and every rejected trajectory contributes nothing.

\subsection{Expected drift without a history-wise supremum}
\label{app:expected-drift}

The local surrogate also admits bounds in terms of expected drift over behavior histories.  The following result retains population expectations in the KL chain-rule argument reviewed by \citet[][Sec.~3.1 and Appendix~B.1]{li2026trustregionmasking}.

\begin{prop}[Expected-history surrogate error]
\label[prop]{prop:expected-drift}
Fix policies $\mu,\pi$, a common prompt distribution and environment, and a fixed horizon $T$ with absorbing padding.  Assume $|A_t^\mu(h,a)|\leq A_{\max}$ on the relevant histories.  Define
\begin{equation}
 e_u=\mathbb E_{h\sim d_\mu^u}[\TV(\mu(\cdot\mid h),\pi(\cdot\mid h))],
 \qquad
 k_u=\mathbb E_{h\sim d_\mu^u}[\KL(\mu(\cdot\mid h)\,\|\,\pi(\cdot\mid h))].
\end{equation}
For the unmasked, unnormalized surrogate in \cref{eq:app-local-surrogate},
\begin{align}
 |J(\pi)-\mathcal L_\mu(\pi)|
 &\leq 2A_{\max}\sum_{u=1}^{T-1}(T-u)e_u,
 \label{eq:expected-tv-drift}\\
 |J(\pi)-\mathcal L_\mu(\pi)|
 &\leq 2A_{\max}\sum_{t=1}^{T}\sqrt{\frac12\sum_{u<t}k_u}.
 \label{eq:expected-kl-drift}
\end{align}
The KL bound is informative when the corresponding divergences are finite.
\end{prop}
\begin{proof}
Let $f_t(h)=\mathbb E_{a\sim\pi(\cdot\mid h)}[A_t^\mu(h,a)]$, so $\|f_t\|_\infty\leq A_{\max}$.  The performance-difference decomposition in \cref{eq:app-pdl} gives
\[
 |J(\pi)-\mathcal L_\mu(\pi)|
 \leq 2A_{\max}\sum_{t=1}^{T}\TV(d_\pi^t,d_\mu^t).
\]
Write $P_\pi^u$ and $P_\mu^u$ for the history-extension kernels, including the shared environment.  Adding and subtracting $d_\mu^uP_\pi^u$, using contraction of TV under a common kernel, and averaging conditional TV under $d_\mu^u$ yields
\[
 \TV(d_\pi^{u+1},d_\mu^{u+1})
 \leq\TV(d_\pi^u,d_\mu^u)+e_u.
\]
The initial history distributions agree, so induction gives $\TV(d_\pi^t,d_\mu^t)\leq\sum_{u<t}e_u$.  Summing proves \cref{eq:expected-tv-drift}.  For complete histories, the KL chain rule gives $\KL(d_\mu^t\|d_\pi^t)=\sum_{u<t}k_u$; common environment transitions contribute zero conditional KL.  Applying Pinsker at each $t$ proves \cref{eq:expected-kl-drift}.
\end{proof}

\section{Optional Negative-Token Filtering}
\label{app:token-mask}

\subsection{Selector and motivation}

Trajectory-level negative advantages also penalize useful intermediate steps in failed rollouts.  Sample-mean reduction addresses length weighting (\cref{sec:reduction-ablation}) but does not identify which steps caused failure.  Here we examine how retaining different fractions of tokens within failed trajectories affects reasoning accuracy and tool use.

\citet{wu2026crf} describe how group-relative advantages allow positive and negative rollouts of the same prompt to partially cancel gradients on shared supporting tokens.  Single-rollout training lacks this within-prompt cancellation.  Sample-mean reduction addresses the separate effect of trajectory length on update weighting.

Without token-level supervision, we rank tokens by policy entropy, following evidence that high-entropy positions carry disproportionate learning signal \citep{wang2025entropy8020}.  Let $F_i=\ind[R_i\leq0]$ denote the failure label in the binary-reward runs, and let $r_{i,t}\in\{1,\ldots,T_i\}$ be the rank of token $t$ by decreasing entropy.  For retention fraction $q\in[0,1]$, define
\begin{equation}
 m_{i,t}^{\mathrm{tok}}
 =\ind\!\left[F_i=0\ \text{or}\ r_{i,t}\leq\lceil qT_i\rceil\right].
 \label{eq:entropy-mask}
\end{equation}
Successful trajectories retain all policy tokens; failed trajectories retain their $\lceil qT_i\rceil$ highest-entropy positions.  The selector is held fixed during the update and multiplies the corresponding token term in \cref{eq:flash-gradient}; it applies to binary-reward settings with an explicit failure label.

\subsection{Entropy diagnostic}

\begin{figure}[H]
\centering
\includegraphics[width=0.88\linewidth]{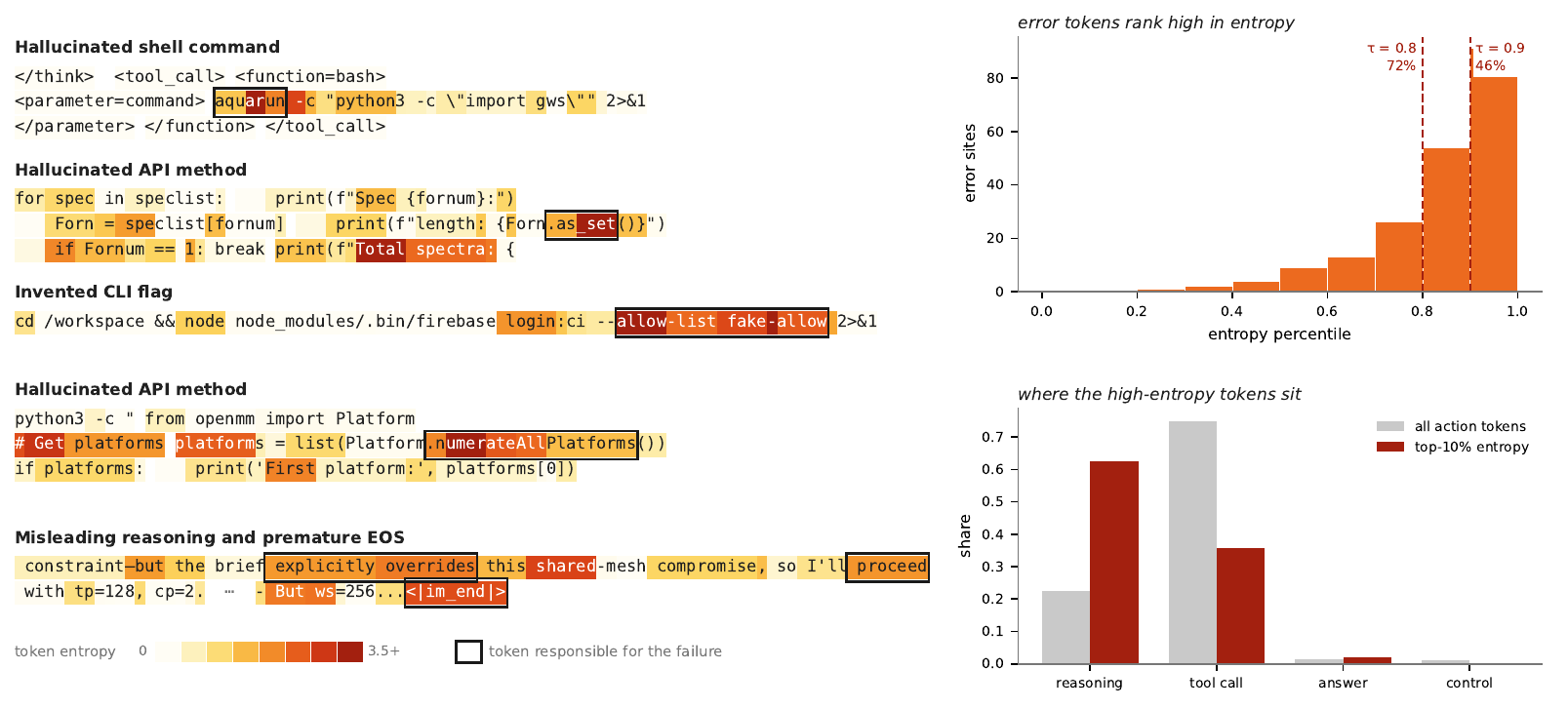}
\caption{\textbf{Entropy diagnostic for the optional selector.}
\textbf{Left:} reward-zero Qwen3.5-4B terminal-task fragments \citep{qwen2026qwen35}; boxes mark failure-causing spans.
\textbf{Right:} 72\% of annotated sites exceed the within-trajectory 80th entropy percentile (versus 20\% for length-matched random spans), and 46\% exceed the 90th.  Top-decile entropy tokens concentrate in reasoning text rather than format-control tokens.}
\label{fig:entropy-failure-analysis}
\end{figure}

\Cref{fig:entropy-failure-analysis} shows that annotated failure sites are concentrated at high-entropy positions, supporting entropy as a ranking heuristic for this diagnostic.

\subsection{Mathematical-reasoning ablation}
\label{app:math-negative-filter}

\begin{table}[H]
\caption{Negative-token filtering on Qwen2.5-Math-1.5B at $\delta=1\times10^{-3}$.  Entries are five-benchmark mean avg@32 accuracy (\%); $q=1$ applies no filtering.  Boldface marks column bests.}
\label{tab:math-token-mask}
\centering\footnotesize\FRtableformat
\begin{tabular}{@{}lrrrrr@{}}
\toprule
\textbf{Setting} & $q$ & \textbf{Step 1,000} & \textbf{Step 2,000} & \textbf{Step 3,000} & \textbf{Step 4,000}\\
\midrule
No filtering & 1.0 & 36.04 & 36.89 & 36.93 & \textbf{37.09}\\
Light entropy filtering & 0.9 & 36.03 & \textbf{37.91} & \textbf{37.40} & 37.01\\
Aggressive entropy filtering & 0.2 & \textbf{37.04} & 36.80 & -- & --\\
\bottomrule
\end{tabular}
\end{table}

Light filtering reaches a peak mean of 37.91 versus 37.09 for the unfiltered reference.  The baseline comparison in \cref{tab:math-main} instead uses $\delta=3\times10^{-3}$ without token filtering.

\subsection{Python tool-use ablation: Qwen2.5-7B-Instruct}
\label{app:tool-negative-filter}

\begin{table}[H]
\caption{Negative-token filtering on Qwen2.5-7B-Instruct Python tool use at $\delta=1\times10^{-3}$.  Best mean is the peak three-benchmark accuracy (\%); $q=1$ applies no filtering.  Tool calls are averaged over steps 601--700.  Boldface marks column bests.}
\label{tab:tool-token-mask}
\centering\footnotesize\FRtableformat
\fittable{%
\begin{tabular}{@{}lrrrr@{}}
\toprule
\textbf{Setting} & $q$ & \textbf{Best mean} & \textbf{Best step} & \textbf{Calls 601--700}\\
\midrule
No filtering & 1.0 & 37.0 & 600 & 3.30\\
Light entropy filtering & 0.9 & \textbf{39.4} & 700 & \textbf{3.72}\\
Aggressive entropy filtering & 0.2 & 32.2 & 200 & 0.03\\
Full failure-token filtering & 0.0 & 13.1 & 100 & 0.00\\
\bottomrule
\end{tabular}}
\end{table}

Light negative-token masking offers a modest advantage.

\end{document}